\documentclass[10pt,twocolumn,letterpaper]{article}

\usepackage{cvpr}              % To produce the CAMERA-READY version
\definecolor{cvprblue}{rgb}{0.21,0.49,0.74}
\usepackage[pagebackref,breaklinks,colorlinks,allcolors=cvprblue]{hyperref}
\usepackage{amsthm}
\usepackage{mathtools}
\usepackage{tikz}
\usepackage{booktabs}
\RequirePackage{colortbl}
\RequirePackage{xcolor}
\RequirePackage{amsmath}
\RequirePackage{amssymb}
\theoremstyle{definition}

\def\confName{CVPR}
\def\confYear{2026}

\title{LSMSeg: Unleashing the Power of Large-Scale Models for Open-Vocabulary Semantic Segmentation}

\author[1]{Huadong Tang}
\author[2]{Youpeng Zhao}
\author[1]{Yan Huang}
\author[1]{Min Xu}
\author[2]{Jun Wang}
\author[1]{Qiang Wu}

\affil[ ]{
\begin{minipage}[t]{0.45\textwidth}
\centering
\textsuperscript{1}University of Technology Sydney
\end{minipage}%
\hspace{-2cm} % 使用负值进一步缩小间距
\begin{minipage}[t]{0.45\textwidth}
\centering
\textsuperscript{2}University of Central Florida
\end{minipage}
}

\affil[ ]{\small\texttt{huadong.tang@student.uts.edu.au, \{yan.huang-7, min.xu, qiang.wu\}@uts.edu.au}}

\affil[ ]{\small\texttt{\{youpeng.zhao, jun.wang\}@ucf.edu}
}

\begin{document}
\maketitle
% \footnotetext[1]{* Corresponding author.}
\begin{abstract}
Open-vocabulary semantic segmentation requires precise pixel-level alignment of visual and textual representations, leveraging text as a universal reference to address visual disparities across diverse datasets. 
While prior efforts have primarily focused on enhancing visual representations or alignment models, the contribution of textual representations remains underexplored. 
Moreover, although CLIP excels at capturing image-level features, its limited capacity for fine-grained pixel-level representation poses a major challenge for semantic segmentation.
To address these challenges, we propose LSMSeg that employs large language models (LLMs) to generate enriched text prompts incorporating diverse visual attributes such as color, shape, size, and texture, thereby replacing simplistic templates with semantically rich descriptions.
In addition, we propose a Feature Refinement Module that adapts visual features from the Segment Anything Model (SAM) to the CLIP space using a lightweight adapter, followed by a learnable weighting strategy to fuse them with CLIP features, enhancing pixel-to-text alignment.
To further reduce computational overhead, we introduce a Category Filtering Module to accelerate training and decrease parameter complexity. 
Extensive experiments demonstrate that LSMSeg significantly enhances cross-modal alignment and achieves strong performance while maintaining efficiency, offering a robust advancement for open-vocabulary semantic segmentation.
\end{abstract}    
\section{Introduction}
Open-vocabulary semantic segmentation (OVSS) seeks to classify each pixel in an image into its most relevant semantic category from a potentially unbounded set, guided by arbitrary or descriptive text inputs~\cite{ovseg,san}.
This task relies heavily on pre-trained vision-language foundation models, such as CLIP~\cite{clip} and Align~\cite{align}, to achieve pixel-level alignment between visual and textual features. 
However, these foundation models trained on image-level paired datasets primarily capture global context rather than localized semantics, limiting their generalization to pixel-level tasks.
\begin{figure}
    % \vspace{-3mm} % 调整与正文的间距
    \centering
    % --- 子图 a ---
    \begin{subfigure}[b]{0.32\linewidth}
        \centering
        \includegraphics[width=2.65cm,height=2.1cm]{figures/image_person.jpg}
        \includegraphics[width=2.65cm,height=2.1cm]{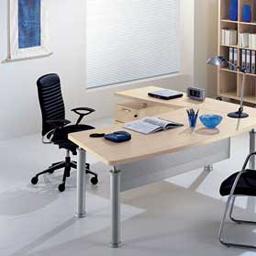}
        \includegraphics[width=2.65cm,height=2.1cm]{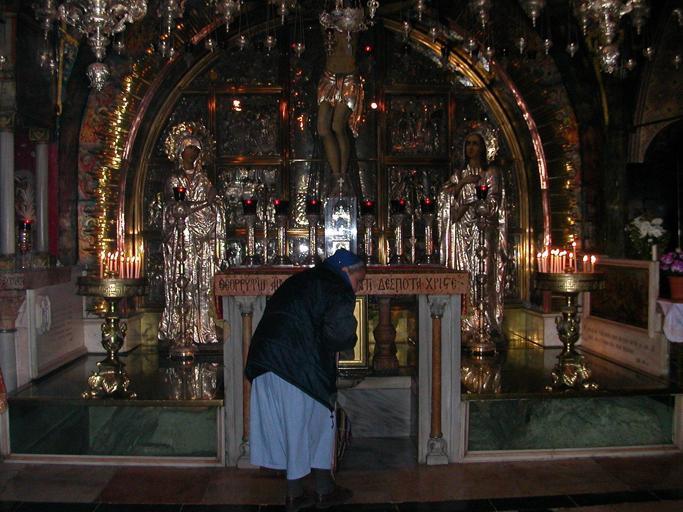}
        \caption{Image}
    \end{subfigure}%
    % --- 子图 b ---
    \begin{subfigure}[b]{0.32\linewidth}
        \centering
        \includegraphics[width=2.65cm,height=2.1cm]{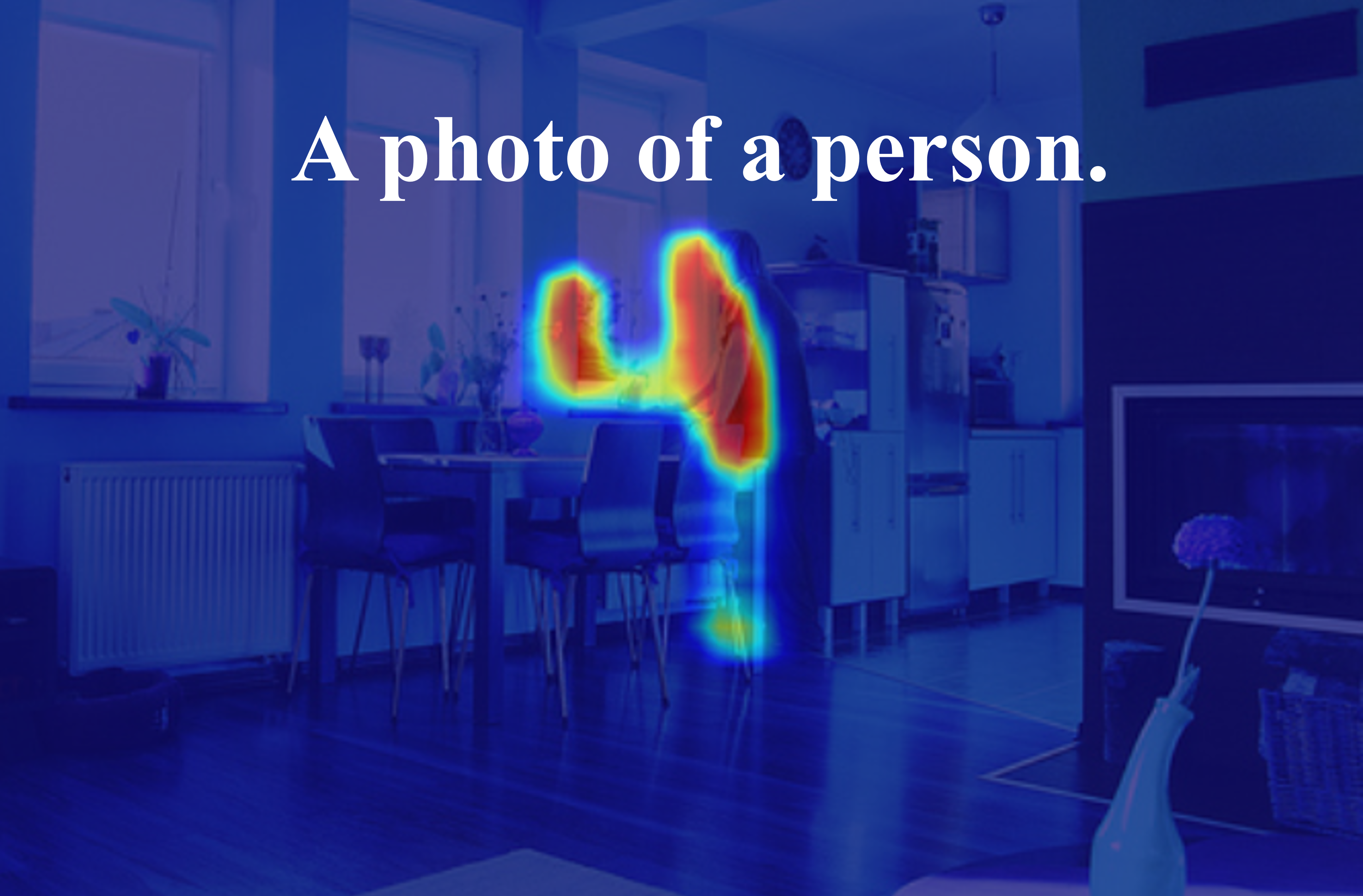}
        \includegraphics[width=2.65cm,height=2.1cm]{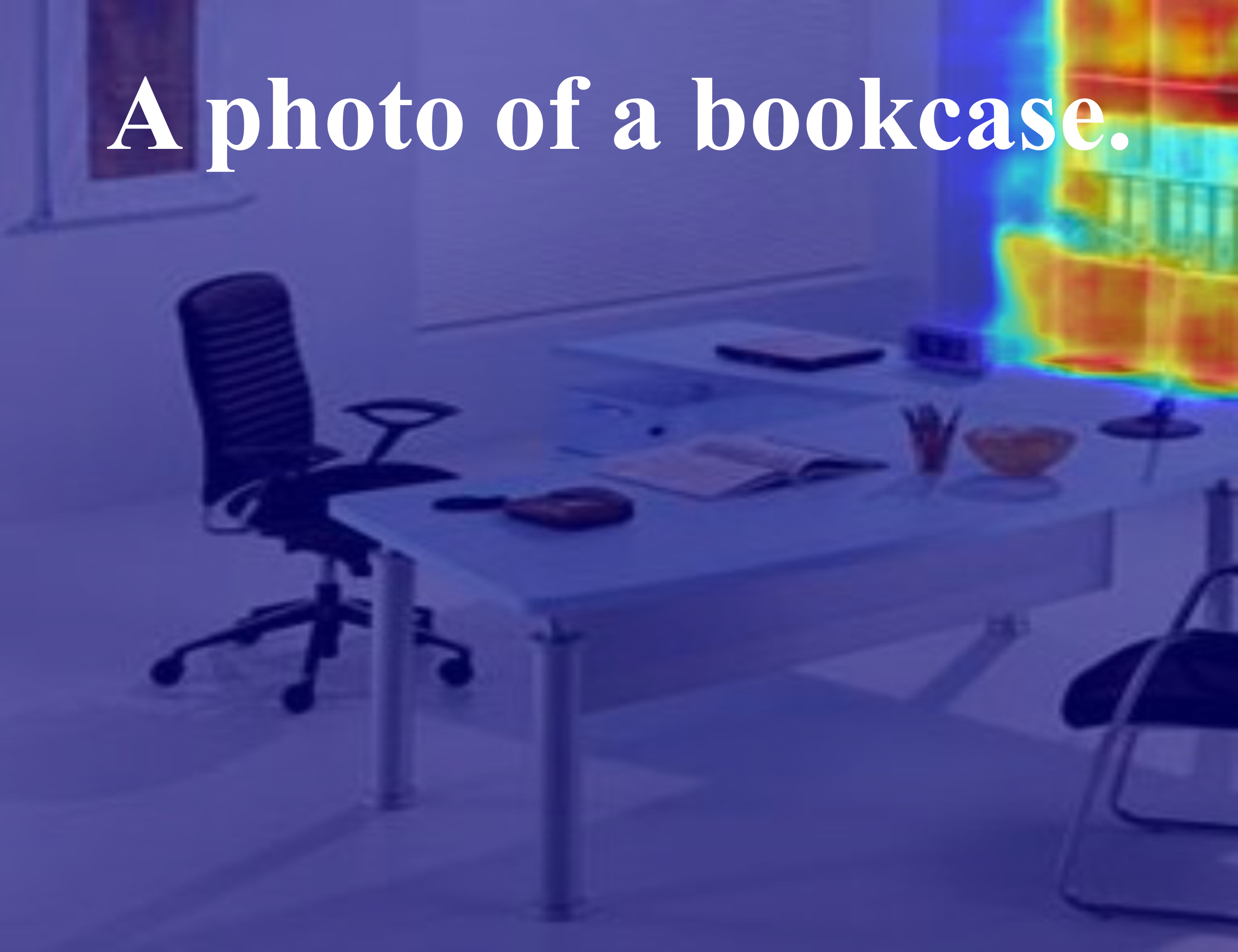}
        \includegraphics[width=2.65cm,height=2.1cm]{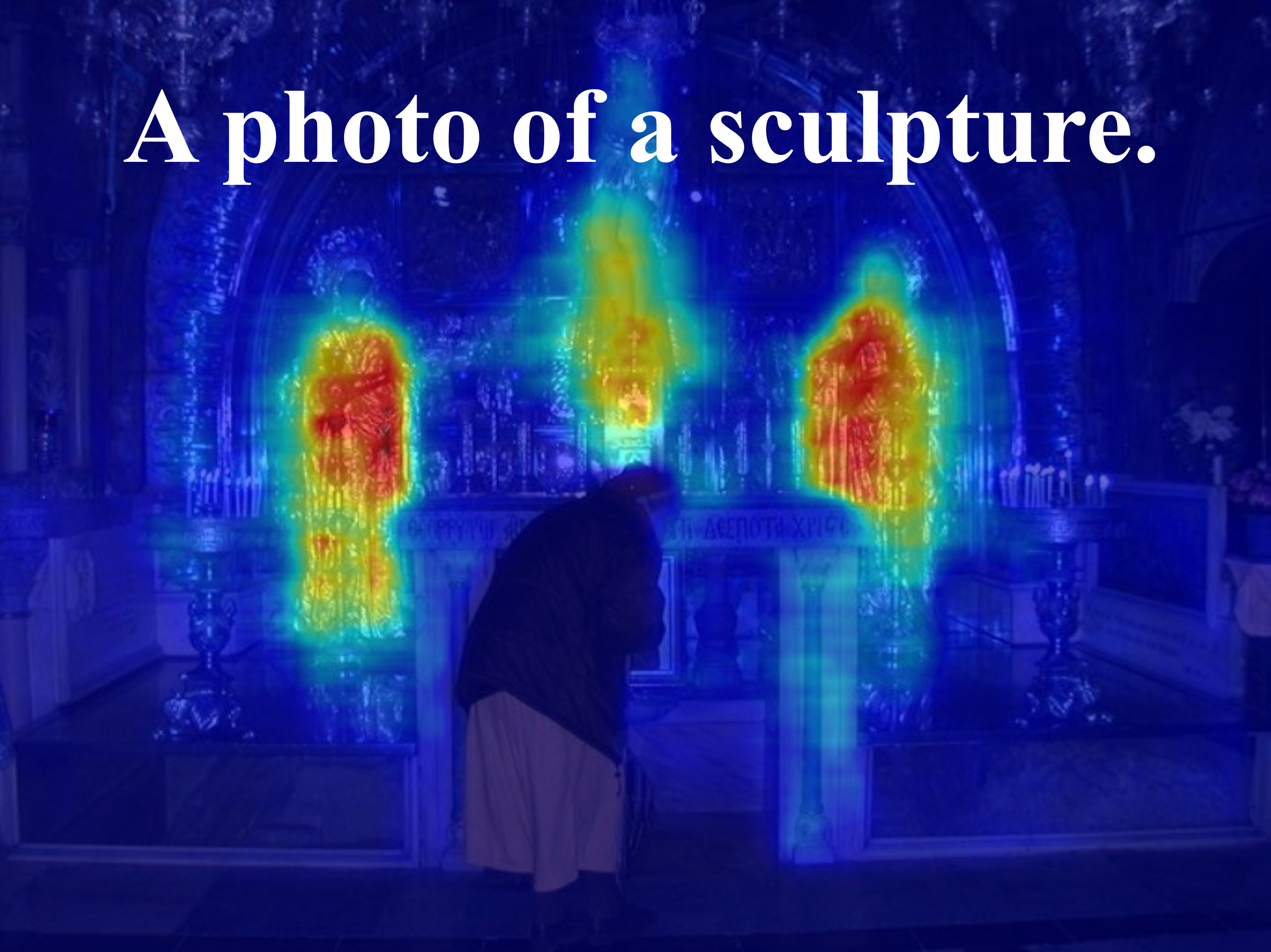}
        \caption{Template Prompt}
    \end{subfigure}%
    % --- 子图 c ---
    \begin{subfigure}[b]{0.32\linewidth}
        \centering
        \includegraphics[width=2.65cm,height=2.1cm]{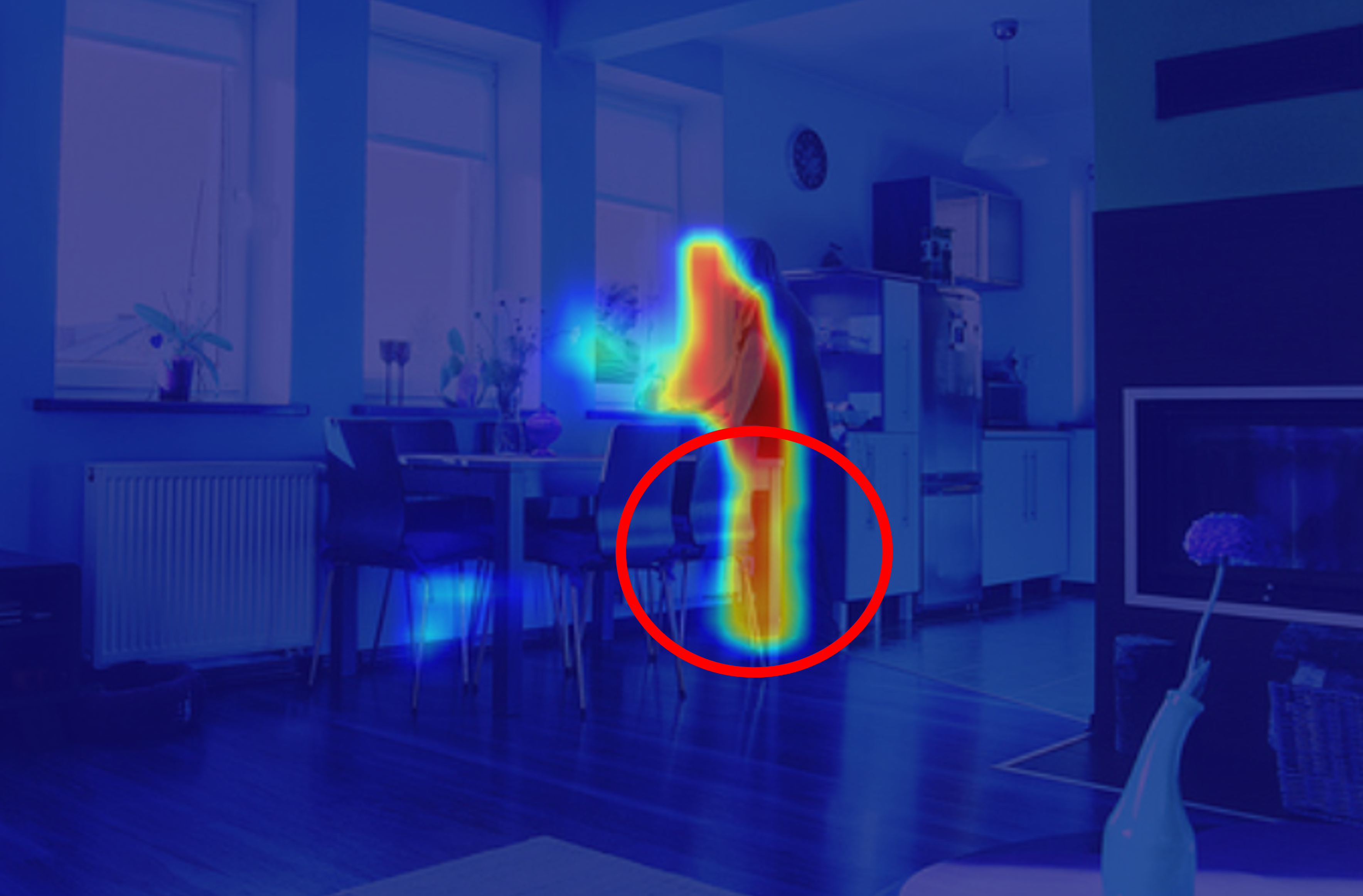}
        \includegraphics[width=2.65cm,height=2.1cm]{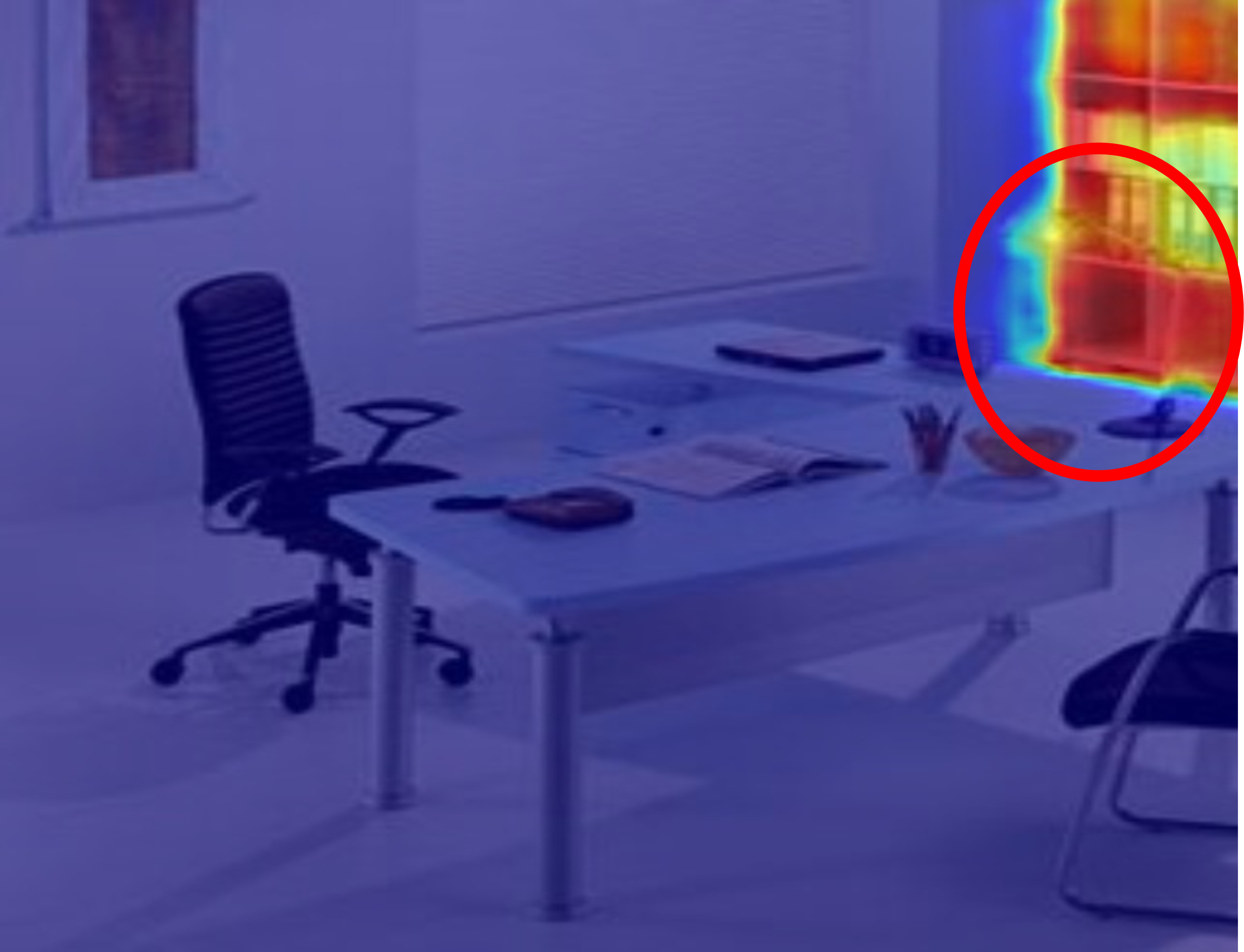}
        \includegraphics[width=2.65cm,height=2.1cm]{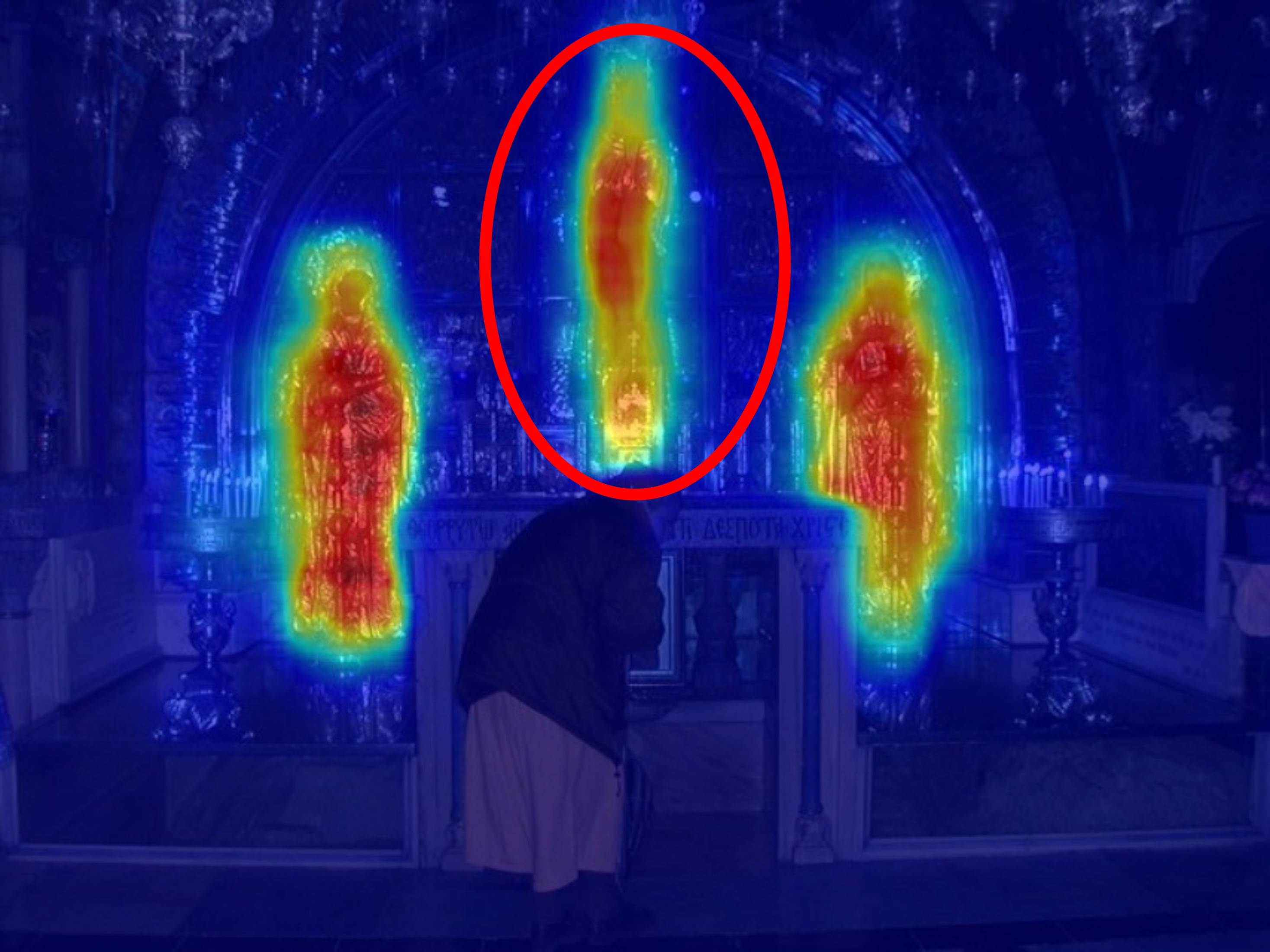}
        \caption{Enriched Prompt}
    \end{subfigure}
    \vspace{-1mm}
    % \captionsetup{width=0.95\linewidth} % 保证 caption 不超出 wrapfigure 宽度
    \caption{\textbf{Visualization of the cost map.} 
    The cost map shows pixel-text alignment, with the first row for the seen class `person' and the last two for the unseen classes `bookcase' and `sculpture'.}
    \label{fig:fig1}
\end{figure}
Existing strategies to improve alignment fall into two main categories: (1) \textbf{Refine region-level visual-text alignment.} 
Methods like~\cite{openseg,ovseg,simseg,maft+} employ a category-agnostic mask generator to derive region-level representations that closely resemble image-level features.
However, these methods primarily achieve region-level alignment, incurring substantial computational costs and inefficient memory usage.
(2) \textbf{Refine pixel-level visual-text alignment.} Other works~\cite{ebseg,sed,catseg} propose to leverage the extra vision foundation models or feature aggregation to enhance rich pixel-level visual representation, thereby compensating for the spatial deficiencies of CLIP features. These methods complement the spatial limitations of CLIP features, which stem from image-level contrastive training that favors global context over local pixel semantics.
Nevertheless, these efforts often overlook a critical component: the quality of textual representations. Simple text prompts like `\texttt{a photo of a \{class name\}}' often lack the \textbf{semantic richness} necessary to resolve fine-grained distinctions. 
Moreover, CLIP’s text encoder can struggle with lexical ambiguities, limiting its discriminative power for fine-grained segmentation tasks.
% However, they primarily concentrate on enhancing visual features while overlooking the critical role of the text prompt.
% These approaches remain limited by fixed, hand-crafted prompts, such as `\texttt{a photo of a dog},' similar to those used in CLIP’s pre-training.
% They are constrained to the fixed hand-crafted prompts like `\texttt{a photo of a dog},' similar to those used during CLIP's pre-training.
% (3) \textbf{Attribute-level visual-text alignment.} Emerging studies have shown that enhancing the quality of text templates by incorporating attributes \cite{yan2023learning, kaul2023multi} or class-specific details \cite{cupl,saha2024improved} can improve CLIP's performance.
% However, these efforts are typically designed for general contexts and seldom customized to meet the specific requirements of open-vocabulary semantic segmentation (OVSS).
% Although existing methods have achieved promising performance, there are still some issues to be addressed: (1) \textbf{Text Prompts}:
This limitation highlights our argument that the quality of textual representations is equally crucial for achieving precise visual-text alignment in OVSS.

Simplistic prompts fall short in three key aspects:
\underline{First}, they lack the detailed semantic information required for fine-grained segmentation tasks, such as differentiating a flower species based on its intricate petal structure and color.
\underline{Second}, the discriminative power of generated text embeddings depends on the CLIP text encoder, which may fail to distinguish between meanings if there are lexical ambiguities. 
For instance, the word `\texttt{bat}' could refer to either `\texttt{a flying mammal}' or `\texttt{a piece of sports equipment used in baseball},' and simply encoding the class name will not be enough to differentiate between these two concepts. 
\underline{Third}, they fail to leverage multi-modal information, which is crucial for capturing the nuances of complex categories, thereby limiting the model's adaptability to diverse and fine-grained visual contexts.
To solve this problem, we propose to \textbf{refine text attribute-level visual-text alignment}. Specifically, we leverage GPT-4 to generate a set of candidate attributes, which are then used to prompt GPT-4 for detailed sentence descriptions tailored to each category.
Compared to simple template-based prompts, the attribute-enriched prompts enable the computation of more informative pixel-to-text cost maps~\cite{catseg}, which capture the fine-grained similarity between textual and visual features. This leads to more accurate cross-modal alignment, as shown in Figure~\ref{fig:fig1}.

\begin{figure}
    \centering
    \vspace{-3mm} 
    \centering
    \includegraphics[width=1\linewidth, keepaspectratio]{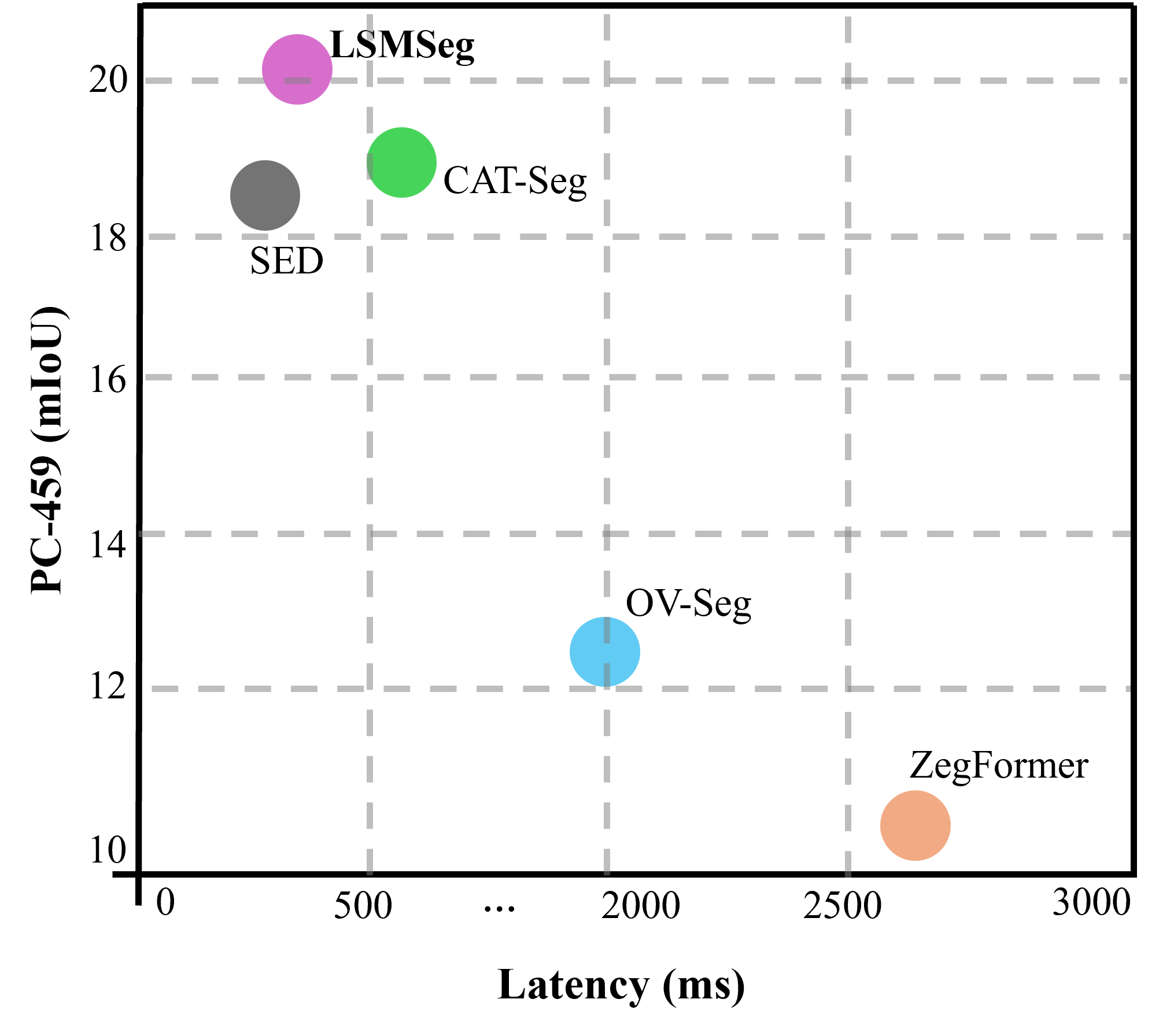}
    \vspace{-5mm}
    % \captionsetup{width=0.9\linewidth}
    \caption{\textbf{Segmentation Performance and inference latency on PC-459.} 
    Our LSMSeg outperforms ZegFormer~\cite{zegformer}, OV-Seg~\cite{ovseg}, 
    CATSeg~\cite{catseg}, and SED~\cite{sed}, achieving a new state-of-the-art 
    mIoU of 20.3\% while maintaining lower latency.}
    \label{fig:la}
\end{figure}

Enhancing textual representations is crucial for capturing attribute-level distinctions in OVSS. However, precise segmentation also depends on high-quality visual features.
Prior works~\cite{GCLIP,cliptrase} have demonstrated that intermediate layers of CLIP effectively capture dense features. 
Motivated by this, we propose a feature refinement module that integrates intermediate CLIP features with a frozen SAM’s image encoder~\cite{sam} and refines the cost map using a Swin-Transformer block and a subsequent linear Transformer block.
Meanwhile, we introduce a category filtering module (CFM) to eliminate irrelevant classes and lower computational complexity. By pruning low-relevance categories from the initial cost map, the module improves segmentation accuracy and efficiency. As illustrated in Figure~\ref{fig:la}, LSMSeg achieves a new state-of-the-art for both efficiency and accuracy.
In summary, our main contributions can be summarized as:(1) We propose LSMSeg, a pioneering framework that leverages large language models (LLMs) to create detailed, attribute-enriched text prompts, significantly improving text-visual alignment for OVSS. (2) We propose a feature refinement module by utilizing the precise spatial information of SAM with a category filtering module to reduce computational cost. (3) Extensive experiments across multiple benchmarks demonstrate that LSMSeg achieves state-of-the-art performance in open-vocabulary semantic segmentation.

\section{Related Work}
\subsection{Open-Vocabulary Semantic Segmentation}
Prevalent semantic segmentation methods~\cite{deeplabv3+,tang2023class,ocrnet, hu2021region, mcibi, mcibi++} are designed for closed sets where only predefined categories can be distinguished, and gathering data for training such models is often costly and time-consuming. 
As a result, the trend in segmentation tasks is shifting towards open-vocabulary approaches.
Existing works primarily focus on two approaches: (1) Refine region-level visual-text alignment.
Some works~\cite{openseg,ovseg,simseg,zegformer} adopt a two-stage framework to refine the alignment, where it first trains a class-agnostic mask generator to extract masks and then leverages the pre-trained CLIP to classify each mask.
% \begin{wrapfigure}{r}{0.4\linewidth} 
% \end{wrapfigure}
OVSeg~\cite{ovseg} proposes fine-tuning the pre-trained CLIP on these images and building a domain-specific training dataset to address the CLIP's recognition ability on masked background regions.
Such a two-stage approach is inefficient and suboptimal, as it uses separate networks for mask generation and classification, lacks contextual information, and incurs high computational costs by requiring CLIP to process multiple image crops. 
(2) Refine pixel-level visual-text alignment.
Unlike two-stage approaches, recent one-stage methods~\cite{odise,san,scan} directly apply a unified vision-language model for open-vocabulary segmentation. 
SAN~\cite{san} attaches a lightweight image encoder to the pre-trained CLIP to generate masks and attention biases. 
SCAN~\cite{scan} proposes a semantic integration module to embed the global semantic understanding and a contextual shift strategy to achieve domain-adapted alignment. 
CATSeg~\cite{catseg} introduces a cost aggregation-based framework, incorporating spatial and class aggregation to reason over the multi-modal cost volume effectively.
Although these methods demonstrate efficacy, they tend to neglect the pivotal role of language in open-vocabulary semantic segmentation, relying solely on extracting text embeddings from pre-trained vision-language models (VLMs), with almost no works focusing on refining text attributes.
\subsection{Text Prompt Enhancement with LLMs}
Recently, numerous large language models (LLMs), such as GPT~\cite{gpt} and LLaMA~\cite{llama}, have been introduced. 
Several works~\cite{cupl,ProText,WaffleCLIP,CoPrompt} have showcased their capability to enhance the performance of early vision-language models (VLMs) with LLMs.
CuPL~\cite{cupl} leverages large language models to produce class-specific prompt descriptions, which are then utilized for text prompt ensembling.
WaffleCLIP~\cite{WaffleCLIP} uses random descriptors and demonstrates additional improvements by incorporating data-specific concepts generated through LLMs.
CoPrompt~\cite{CoPrompt} harnesses a pre-trained large language model's expertise, applying coherence constraints to the text component and data enhancement to the image component to further improve generalization.
Some methods~\cite{lai2024lisa} leverage LLMs for reasoning segmentation, highlighting their promise in detailed visual understanding.
In this work, we design LLM-generated prompts for OVSS by optimizing attribute selection and combination at the pixel level. This task-specific improvement distinguishes our approach, addressing the unique challenge of aligning fine-grained visual and textual data in segmentation.
\begin{figure*}[]
\begin{center}
% \fbox{\rule{0pt}{2in} \rule{.9\linewidth}{0pt}}
\includegraphics[width=1\textwidth, keepaspectratio]{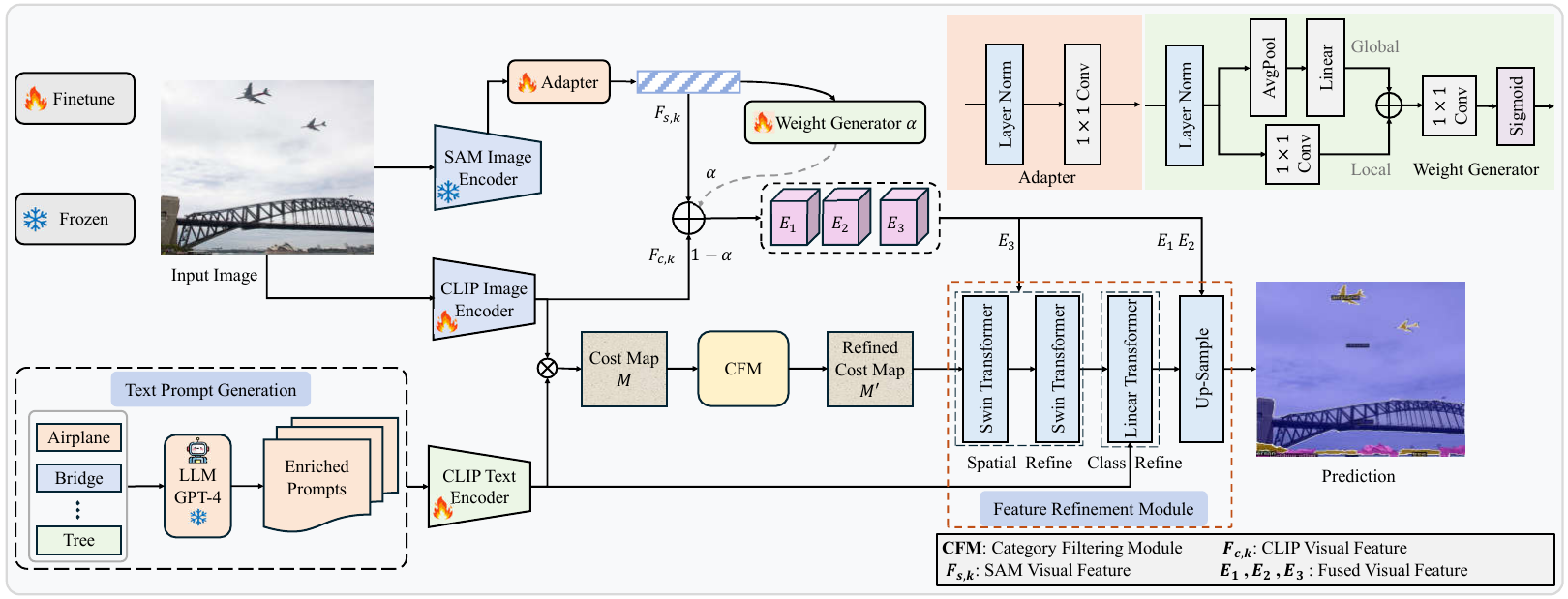}
\end{center}
% \vspace{-2mm}
\caption{\textbf{Overall architecture of our proposed LSMSeg.} We first utilize GPT-4 to generate enhanced text prompts. Next, we propose a category filter module to eliminate irrelevant classes, yielding a refined cost map and reducing computational complexity. Finally, we leverage SAM visual features to address the spatial information deficiency in CLIP visual features through a learnable adapter and weight generator, followed by a feature refinement process to enhance the filtered cost map at both spatial and class levels.}
\label{fig:arc}
\end{figure*}
\section{Method}
\subsection{Preliminaries}
\label{sec1}
\paragraph{Problem Definition.} 
Open-vocabulary semantic segmentation aims to partition an image $I\in \mathbb{R}^{H\times W\times 3}$ into distinct semantic regions based on text descriptions, including classes that were not seen during training.
In training, only pixel-level annotations of the seen classes $C_{train}$ are used, with knowledge of their existence and quantity (\emph{i.e.}, how many and what classes are present).
Annotations for unseen categories are replaced with an “ignored” label.
In an open-vocabulary setting, $C_{test}$ may include new categories that were not encountered during training, meaning $C_{train} \neq C_{test}$.
In inference, both seen and unseen classes need to be segmented.

\subsection{Architecture Overview}
Figure~\ref{fig:arc} illustrates the overall architecture of our proposed LSMSeg, comprising three main components: 
(a) The \textit{Text Prompts Generation} leverages the GPT-4 model to first select appropriate attributes and then generate descriptive sentences based on those attributes, which are subsequently processed by the CLIP text encoder to obtain text features.
% (b) The \textit{Visual Feature Fusion} integrates SAM features with CLIP visual features through a learnable weighted fusion strategy to enhance spatial information representation.
(b) The \textit{Category Filter Module} is proposed to reduce computational parameters and accelerate training by filtering irrelevant classes on the pixel-to-text cost map.
(c) The \textit{Feature Refinement Module } is introduced to integrate SAM features with CLIP visual features through a learnable weighted fusion strategy to enhance spatial information representation. 
This fused representation is utilized to refine spatial-level and class-level feature information of the cost map, enabling a more precise and comprehensive feature representation that improves overall model performance.

\subsection{Text Prompts Generation}
% \begin{wrapfigure}{r}{0.625\textwidth} 
\begin{figure}
\begin{center}
% \fbox{\rule{0pt}{2in} \rule{.9\linewidth}{0pt}}
\includegraphics[width=1\linewidth, keepaspectratio]{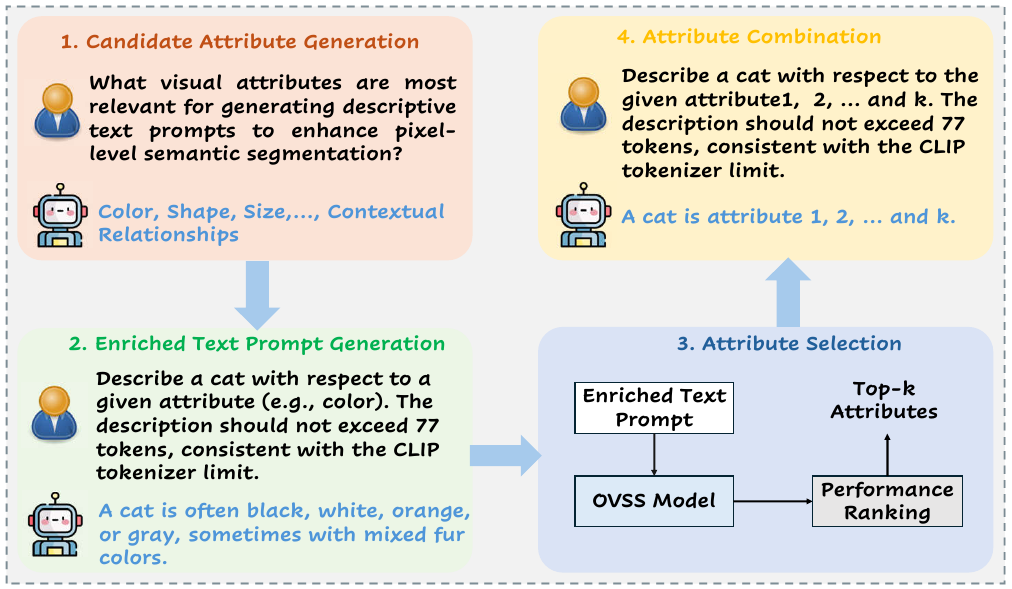}
\end{center}
\vspace{-2mm}
\caption{\textbf{Comprehensive Linguistic Prompt Generation Pipeline.} The pipeline includes four steps: (1) Candidate Attribute Generation; (2) Enriched Text Prompt Generation; (3) Attribute Selection; and (4) Attribute Combination.} 
\label{fig:llm}
% \end{figure*}
\end{figure}
% \end{wrapfigure}
To advance open-vocabulary semantic segmentation (OVSS), we introduce a novel approach that leverages GPT-4 to generate enriched text prompts, thereby enhancing pixel-level alignment between textual and visual features. The overall pipeline is illustrated in Figure~\ref{fig:llm}. Firstly, we query GPT-4 with: `\texttt{What visual attributes are most relevant for generating descriptive text prompts to enhance pixel-level semantic segmentation?}.' This yields nine key attributes: color, shape, size, texture, material, position, pattern, action/state, and contextual relationships.
Secondly, we further prompt GPT-4 with: `\texttt{Describe a \{class name\} with respect to a given \{attribute\}. The description should not exceed 77 tokens, consistent with the CLIP tokenizer limit.}' For instance, given the class `cat' and attribute `color', GPT-4 generates: `\texttt{A cat is often black, white, orange, or gray, sometimes with mixed fur colors.}' This procedure yields fine-grained, attribute-specific descriptions to replace simplistic templates such as `\texttt{a photo of a \{class name\}}', providing richer textual inputs for the CLIP text encoder.
Thirdly, we optimize these prompts by independently assessing the contribution of each attribute without incorporating the Feature Refinement Module.
Finally, we integrate the top-k attributes into comprehensive prompts, such as: `\texttt{A cat has a small, sleek, and agile shape with a long tail and pointed ears, is small to medium-sized, weighing between 3 to 7 kg, has soft, fluffy fur with a smooth or slightly rough tongue, and is often black, white, orange, or gray.}' 

\subsection{Category Filtering Module (CFM)}
Given an input image $I$, we obtain dense visual features $F_{c}\in \mathbb{R}^{B\times H\times W\times D}$ from the CLIP image encoder, where $B$, $H$, $W$, and $D$ represent the batch size, height, width and channel.
Given a set of class names $C$, we leverage LLMs to generate comprehensive linguistic prompts.
We obtain text embeddings $T\in \mathbb{R}^{B\times T\times D}$ by feeding these prompts into the CLIP text encoder, where $T$ and $D$ represent the number of class and channel.
By computing the cosine similarity between the visual feature $E$ and text embedding $T$, we derive the cost map embedding $M$ as:
\begin{equation}
M_{(i,j,n)} = \frac{F_{c}{(i,j)} \cdot T_n}{\|F_{c}{(i,j)}\| \|T_n\|}.
\end{equation}
where $i,j$ denotes the spatial positions, and $n$ indicates the text embedding index. Thus, the cost map embedding $M$ has the dimension of $B\times T\times d\times H\times W$, where $d$ is the channel of the cost map embedding.

To reduce computational overhead and suppress noisy or uninformative text tokens, we apply a top-k token selection when the number of text tokens exceeds a predefined padding threshold $q$. Specifically, we compute the maximum correlation across spatial dimensions and visual prompts:
\begin{equation}
\mathbf{A} = \max_{h, w, d} \left( \mathbf{M} \right), \quad \mathbf{A} \in \mathbb{R}^{B \times T},
\end{equation}
Then, we select the indices of the top-$k$ highest responding tokens:
\begin{equation}
\mathcal{I}_k = \texttt{TopK}(\mathbf{A},\ k = q).
\end{equation}
These selected token embeddings are gathered and re-normalized:
\begin{equation}
\mathbf{T}' = \texttt{Gather}(\texttt{Norm}(\mathbf{T}),\ \mathcal{I}_k), \quad \mathbf{T}' \in \mathbb{R}^{B \times k \times D}
\end{equation}

where $\texttt{Norm}(\cdot)$ denotes $\ell_2$ normalization along the feature dimension. $\texttt{Gather}(\cdot, \mathcal{I})$ retrieves the top-$k$ token embeddings along the token dimension based on the selected indices $\mathcal{I}_k$. For a tensor $\mathbf{T} \in \mathbb{R}^{B \times T \times D}$ and index set $\mathcal{I}_k \in \mathbb{R}^{B \times k}$, this operation selects $k$ tokens per sample in the batch and preserves the prompt and feature structure. Then, we recompute the refined cross-modal cost map via:
\begin{equation}
M'_{(i,j,n)} = \frac{F_{c}{(i,j)} \cdot T'_n}{\|F_{c}{(i,j)}\| \|T'_n\|}.
\end{equation}

% \begin{wrapfigure}{r}{0.4\textwidth} % r=右侧, l=左侧, 宽度自己调
%     \vspace{-6mm} % 调整与上文距离
%     \centering
%     \includegraphics[width=0.9\linewidth, keepaspectratio]{figures/FRM.pdf}
%     \vspace{-3mm}
%     % \captionsetup{width=0.9\linewidth} % 让 caption 不超出图宽
%     \caption{\textbf{The Feature Refinement Module.} 
%     We first perform spatial-level feature enhancement and then 
%     aggregate class-level features.}
%     \label{fig:fe}
% \end{wrapfigure}
\subsection{Feature Refinement Module}
\label{sec5}
The CLIP~\cite{clip} model is trained by image-level contrastive learning and struggles with precisely localizing pixel-level visual features.
Its embeddings focus on global visual context instead of the pixel-level semantics within the image. This can be a problem for segmentation, which requires understanding the local context of each pixel about its neighbors.
To address this, we further propose leveraging a frozen SAM~\cite{sam} image encoder to enhance and supplement the spatial information. 
As shown in Figure~\ref{fig:arc}, we input image $I$ into the SAM image encoder and extract image features $F_{s}\in \mathbb{R}^{B\times H\times W\times D_s}$ from the last three global attention blocks. 
A lightweight adapter is proposed to project the SAM features into the same dimensional space as the CLIP features.
Next, the Weight Generator generates an adaptive weighting coefficients through a local and global branch in Figure~\ref{fig:arc}.
This weight control the relative contributions of CLIP and SAM features during fusion:
\begin{equation}
E_k = \alpha\times F_{c,k} + (1-\alpha)\times F_{s,k}
\end{equation}
where $\alpha$ is the weight generator, k stands for different layers and $E$ is the fused visual feature. 

As a segmentation task, it is intuitive to further explore spatial-level and class-level information.
We first utilize the Swin-Transformer block~\cite{liu2021swin} to process the fused visual features for enriching spatial feature information as in~\cite{catseg,sed}. 
Then, we perform class-level feature refinement to map textual information onto each pixel, achieving more precise alignment. 
We feed the text embedding into a linear transformer block generated from the comprehensive prompts through the CLIP text encoder. 
Finally, we leverage the fused feature again to up-sample the enhanced feature representations. The overall process of feature refinement is summarized as follows:
% \vspace{-2mm}
\begin{equation}
M''_{(i,j,n)} = S\left([M'_{(i,j,n)}; E_k]\right),
\end{equation}
\begin{equation}
M'''_{(i,j,n)} = C\left([M''_{(i,j,n)}; T'_n]\right),
\end{equation}
\begin{equation}
O = Up\left([M'''_{(i,j,n)}; E_k]\right),
\end{equation}

% where $S$ and $C$ denote spatial-level and class-level refinement. $F'$ denotes the intermediate layer visual feature.$ Up$ represents the up-sampling operation and $O$ is the final prediction.
where $S$ and $C$ represent spatial-level and class-level refinements, $E'$ is the intermediate visual feature layer, $Up$ denotes up-sampling, and $O$ is the final prediction.

\section{Experiments}
\subsection{Dataset and evaluation protocol}
We train our model on COCO Stuff~\cite{coco} dataset, and conduct evaluation on ADE20k-847~\cite{ade20k} ADE20k-150~\cite{ade20k}, Pascal Context-459~\cite{pc}, Pascal Context-59~\cite{pc}, and Pascal VOC~\cite{voc}.
COCO-Stuff dataset contains 171 annotated classes and includes 118k training, 5k validation, and 41k test images. 
{ADE20K}~\cite{ade20k} is a large-scale benchmark for scene understanding, comprising 20k training images, 2k validation images, and 3k testing images. It includes two sets of annotated classes: ADE20K-150 with 150 classes and ADE20K-847 with 847 classes, although both use the same images. 
{Pascal Context}~\cite{pc} extends Pascal VOC 2010, offering 4,998 training and 5,005 validation images, with annotations available in two configurations: PC-59 (59 classes) and PC-459 (459 classes).
{Pascal VOC}~\cite{pc} consists of 11,185 training images and 1,449 validation images across 20 object classes.
Following previous work~\cite{catseg,fc-clip,san}, Mean Intersection over Union (mIoU) is used as the evaluation metric across all experiments. This metric represents the average intersection-over-union values calculated for each class across all classes.
\subsection{Implementation Details}
Our experiments utilize the pre-trained CLIP model from OpenAI~\cite{clip}, specifically the ViT-B/16 and ViT-L/14 variants. 
We fine-tune the CLIP image and text encoder and the total training iteration is set as 80k.
The initial learnable fusion weight is empirically set as 0.5 for balance.
We use 2 NVIDIA-L40 GPUs for training with a batch size of 4 and the AdamW optimizer with an initial learning rate of $2 \times 10^{-4}$.
The weight decay is $1 \times 10^{-4}$ for our model.
During training, the input image resolution is $384\times 384$ for ViT-B/16. For ViT-L/14, the resolution is $336\times 336$.
\subsection{Comparisons with State-of-the-art Methods}
\begin{table*}[h]
\renewcommand\arraystretch{1}
\centering
\resizebox{0.98\textwidth}{!}{
{
% \small 
% \normalsize
\begin{tabular}{c|c|c|c|c|c|c|c|c} \hline
\toprule
Method & VLM & Training Dataset & A-847 & PC-459 & A-150 & PC-59 & VOC &VOCb \\ \hline
SPNet~\cite{spnet} & - & PASCAL VOC & - & - & - & 24.3 & 18.3 & - \\
ZS3Net~\cite{zs3net} & - & PASCAL VOC & - & - & - & 19.4 & 38.3 & - \\\hline
Lseg+~\cite{lseg} & ALIGN EN-B7 & COCO-Stuff & 3.8 & 7.8 & 18.0 & 46.5 & - & - \\
OpenSeg~\cite{openseg} & ALIGN EN-B7 & COCO Panoptic & 8.1 & 11.5 & 26.4 & 44.8 & - & 70.2\\ \hline
ZegFormer~\cite{zegformer} & CLIP ViT-B/16 & COCO-Stuff & 5.6 & 10.4 & 18.0 & 45.5 & 89.5 & 65.5\\
DeOP~\cite{deop} & CLIP ViT-B/16 & COCO-Stuff-156 & 7.1 & 9.4 & 22.9 & 48.8 & 91.7 & -\\
% PACL~\cite{pacl} & CLIP ViT-B/16 & GCC + YFCC & 7.1 & 9.4 & 22.9 & 48.8 & 91.7 & -\\
OVSeg~\cite{ovseg} & CLIP ViT-B/16 & COCO-Stuff & 7.1 & 11.0 & 24.8 & 53.3 & 92.6 & -\\
SAN~\cite{san} & CLIP ViT-B/16 & COCO-Stuff & 10.1 & 12.6 & 27.5 & 53.8 & 94.0 & -\\
SCAN~\cite{scan} & CLIP ViT-B/16 & COCO-Stuff & 10.8  & 13.2 & 30.8 & 58.4 & \textbf{97.0} & -\\
EBSeg~\cite{ebseg} & CLIP ViT-B/16 & COCO-Stuff & 11.1 & 17.3 & 30.0 & 56.7 & 94.6 & -\\
SED~\cite{sed} & ConvNeXt-B & COCO-Stuff & 11.4 & 18.6 & 31.6 & 57.3 & 94.4 & -\\
CAT-Seg~\cite{catseg} & CLIP ViT-B/16 & COCO-Stuff & \underline{12.0} & \underline{19.0} & \underline{31.8} & \underline{57.5} & 94.6 & \underline{77.3}\\\hline
 \rowcolor{darkgray!20}
 LSMSeg (\emph{ours}) & CLIP ViT-B/16 & COCO-Stuff & \textbf{13.1} & \textbf{20.3} & \textbf{33.2} & \textbf{59.7} & \underline{95.4} & \textbf{81.1}\\\hline
 SimSeg~\cite{simseg} & CLIP ViT-L/14 & COCO-Stuff & 7.1 & 10.2 & 21.7 & 52.2 & 92.3 & -\\
 % MaskCLIP~\cite{maskclip} & CLIP ViT-L/14 & COCO Panoptic & 8.2 & 10.0 & 23.7 & 45.9 & - & -\\
 OVSeg~\cite{ovseg} & CLIP ViT-L/14 & COCO-Stuff & 9.0 & 12.4 & 29.6 & 55.7 & 94.5 & -\\
 ODISE~\cite{odise} & CLIP ViT-L/14 & COCO-Stuff & 11.1 & 14.5 & 29.9 & 57.3 & -& -\\
 SAN~\cite{san} & CLIP ViT-L/14 & COCO-Stuff & 12.4 & 15.7 & 32.1 & 57.7 & 94.6 & -\\
 EBSeg~\cite{ebseg} & CLIP ViT-L/14 & COCO-Stuff & 13.7 & 21.0 & 32.8 & 60.2 & 96.4 & -\\
 SCAN~\cite{scan} & CLIP ViT-L/14 & COCO-Stuff & 14.0  & 16.7 & 33.5 & 59.3 & 97.2 & -\\
 FC-CLIP~\cite{fc-clip} & ConvNeXt-L & COCO Panoptic & 14.8 & 18.2 & 34.1 & 58.4 & 95.4 & 81.8\\
 SED~\cite{sed} & ConvNeXt-L & COCO-Stuff & 13.9 & 22.6 & 35.2 & 60.6 & 96.1 & -\\
MAFT+~\cite{maft+} & CLIP ViT-L/14 & COCO-Stuff & 15.1& 21.6 & 36.1 & 59.4 & 96.5 & -\\
DPSeg\cite{DPSeg} & ConvNeXt-L & COCO-Stuff & 14.9& 23.5 & 36.4 & 62.0 &  \textbf{97.4} & -\\
CAT-Seg~\cite{catseg} & CLIP ViT-L/14 & COCO-Stuff & 16.0& \underline{23.8} & 37.9 & \underline{63.3} & 97.0 & \underline{82.5}\\
MaskAdapter~\cite{maskadapter} & ConvNeXt-L & COCO-Stuff & \underline{16.2}& 22.7 & \underline{38.2} & 60.4 & 95.8 & -\\\hline
\rowcolor{darkgray!20}
LSMSeg (\emph{ours}) & CLIP ViT-L/14 & COCO-Stuff & \textbf{16.9} & \textbf{25.6} & \textbf{38.5} & \textbf{63.4} & \underline{97.2} & \textbf{84.0}\\
\bottomrule
\end{tabular}
}}
\vspace{-1mm}
\caption{\textbf{Comparison with state-of-the-art methods.} We present the mIoU results on six commonly used test sets for open-vocabulary semantic segmentation. The highest results are highlighted in bold, and the second highest are underlined. Compared with other methods, our proposed LSMSeg demonstrates superior performance across all six test sets.} \label{tab1}
\end{table*}

We compare our method with existing state-of-the-art approaches across six datasets in Table~\ref{tab1}, including the vision-language model (VLM) and training dataset.
Apart from SPNet~\cite{spnet} and ZS3Net~\cite{zs3net}, most methods are developed using VLM as a foundation.
To ensure a fair comparison, the results using the same vision-language model are grouped together.
% \begin{wrapfigure}{r}{0.6\textwidth}
\begin{figure*}
% \vspace{-2mm} % 与上文距离
\centering
% 第一列：原图
\begin{subfigure}[b]{0.24\textwidth}
    \centering
     \includegraphics[width=1\textwidth,height=3cm]{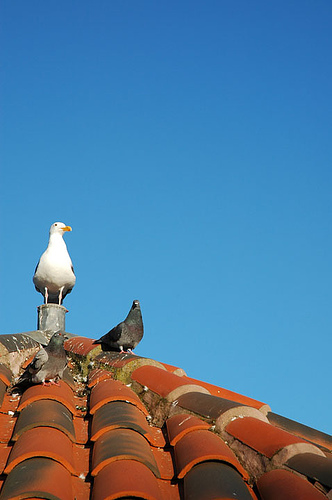}
     \includegraphics[width=1\textwidth,height=3cm]{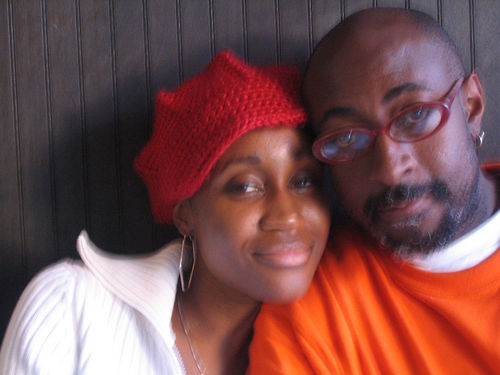}
     \includegraphics[width=1\textwidth,height=3cm]{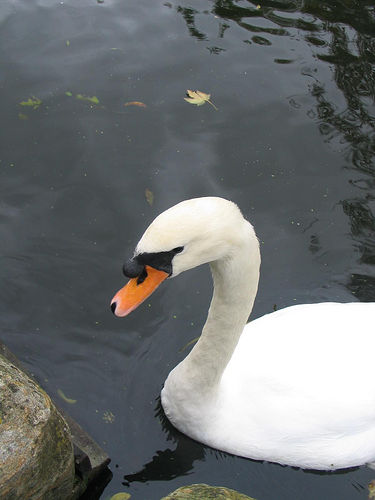}
    \caption{Image}
\end{subfigure}
% 第二列：CAT-Seg
\begin{subfigure}[b]{0.24\textwidth}
    \centering
     \includegraphics[width=1\textwidth,height=3cm]{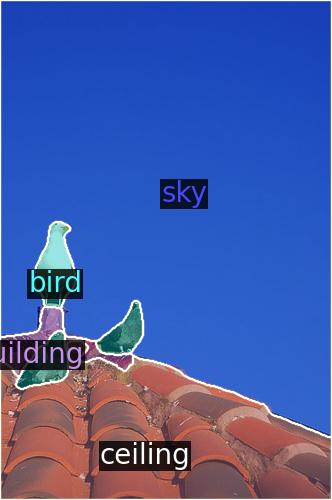}
     \includegraphics[width=1\textwidth,height=3cm]{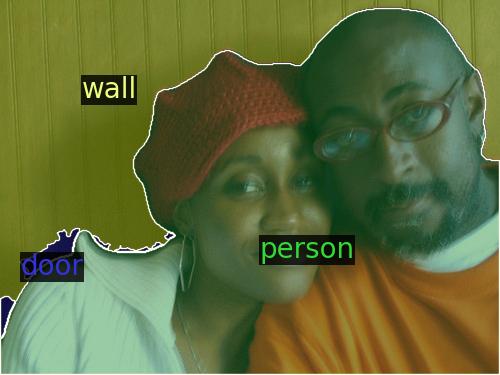}
     \includegraphics[width=1\textwidth,height=3cm]{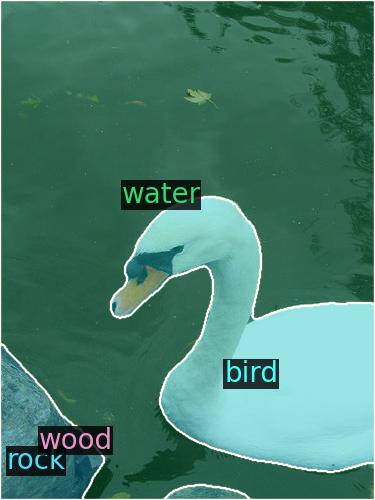}
    \caption{CATSeg}
\end{subfigure}
% 第三列：LSMSeg
\begin{subfigure}[b]{0.24\textwidth}
    \centering
     \includegraphics[width=1\textwidth,height=3cm]{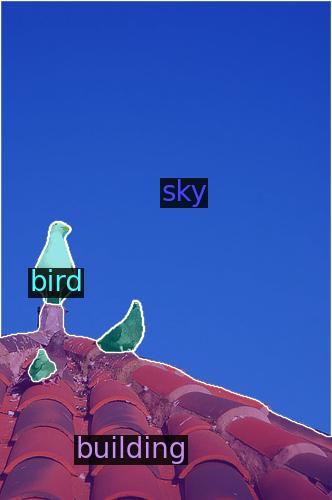}
    \includegraphics[width=1\textwidth,height=3cm]{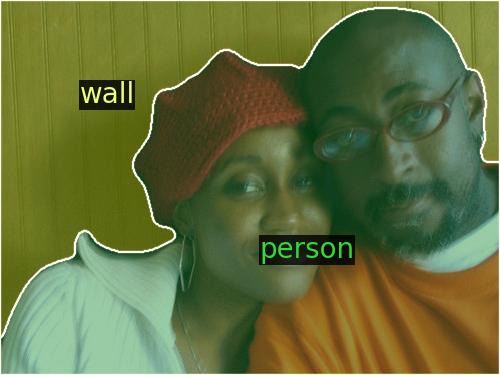}
     \includegraphics[width=1\textwidth,height=3cm]{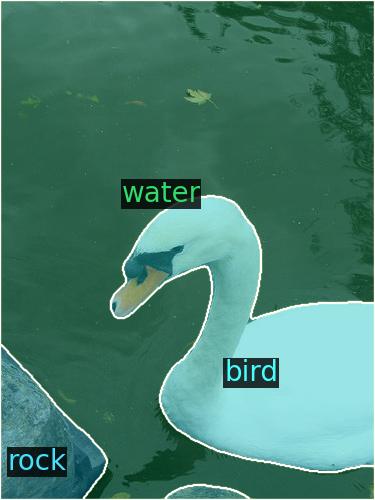}
    \caption{LSMSeg (\emph{ours})}
\end{subfigure}
% 第四列：GT
\begin{subfigure}[b]{0.24\textwidth}
    \centering
    \includegraphics[width=1\textwidth,height=3cm]{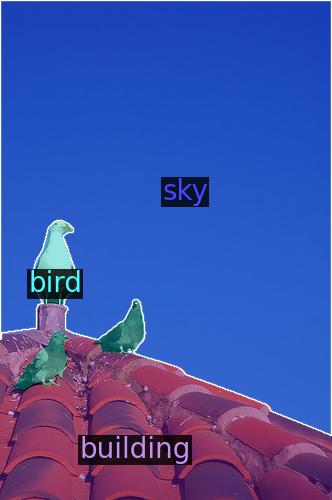}
     \includegraphics[width=1\textwidth,height=3cm]{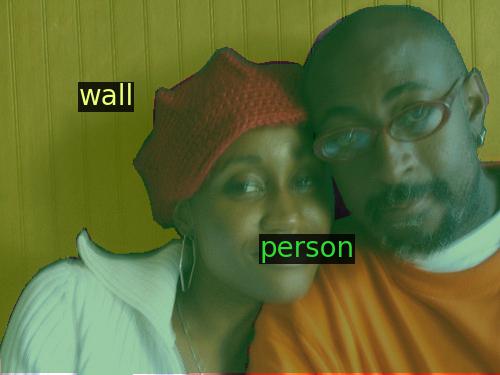}
     \includegraphics[width=1\textwidth,height=3cm]{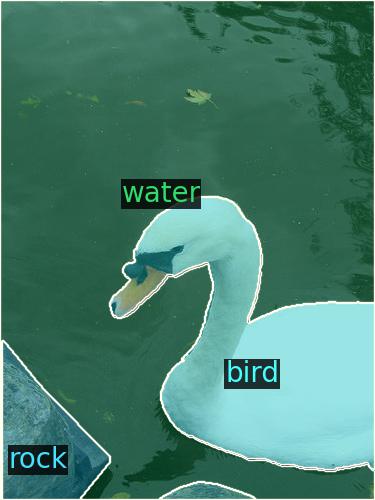}
    \caption{GT}
\end{subfigure}
\vspace{-1mm}
\caption{\textbf{Qualitative comparisons on PC-59.} From left to right: input images, results of CAT-Seg, results of our LSMSeg, and ground truth.}
\label{fig:vis}
\end{figure*}
Existing open-vocabulary semantic segmentation methods explore various vision–language strategies but still struggle to segment unseen classes accurately.
In contrast, our approach achieves notable performance in accurately segmenting both seen and unseen classes.
With ViT-B/16 as the vision-language model, our LSMSeg outperforms SAN~\cite{san}, SED~\cite{sed}, and CATSeg~\cite{catseg} by 5.7\%, 1.6\%, and 1.4\% on A-150. On PC-459, our method exceed SED~\cite{sed}, EBSeg~\cite{ebseg}, and CAT-Seg~\cite{catseg} by 1.7\%, 3.0\%, and 1.3\%.
When using a larger model ViT-L, our LSMSeg also attains notable performance on all six datasets. 
For instance, on ADE-150, our LSMSeg outperforms FC-CLIP~\cite{fc-clip}, SED~\cite{sed} and CAT-Seg~\cite{catseg} by 4.4\%, 3.3\% and 0.6\%. 
Our method achieves favorable performance with both base and large models.
Additionally, we present qualitative comparisons on PC-59 in Figure~\ref{fig:vis}, demonstrating the superior effectiveness of our proposed LSMSeg approach relative to the cutting-edge method. 
% In Figure~\ref{fig:fig1}, our method shows a more focused and accurate activation, highlighting its effectiveness in distinguishing both seen and unseen classes while also demonstrating improved pixel-text alignment.
\subsection{Ablation Studies}
\label{sec4.4}
\paragraph{Analysis of different prompts.} 
We have obtained different visual attributes such as color, shape, size, texture, material, position or location, pattern, action or state, and contextual relationship. 
Here, we utilize ViT-B/16 as the VLM and train on the COCO-Stuff dataset without the feature refinement module.
To identify the attributes that play the most crucial role in the segmentation task, we carried out an extensive series of experiments, with the results succinctly summarized in Table~\ref{tab2_}.
% \begin{wraptable}{r}{0.6\textwidth}
\begin{table}
% \vspace{-3mm}
% \renewcommand\arraystretch{0.5}
% \setlength{\tabcolsep}{0.8mm}
\centering
\renewcommand\arraystretch{1.3}
\resizebox{1\linewidth}{!}{
% {
% \small 
\begin{tabular}{c|c|c|c|c|c|c|c} \hline
\toprule
Methods  & A-847 & PC-459 & A-150 & PC-59 & VOC & VOCb & avg. \\ \hline
Baseline & 11.0 & 17.9 & 28.4 & 54.6 & 94.6 & 74.3 & 46.8\\
\rowcolor{darkgray!20}
Color  & 11.2 & 17.8 & 28.3 & 55.3 & 94.5 & 76.5 & 47.3\\
\rowcolor{darkgray!20}
Shape & 11.2 & 18.4 & 28.2 & 54.7 & 94.9 & 77.1 & 47.4\\
\rowcolor{darkgray!20}
Size & 11.6 & 18.2 & 28.3 & 55.8 & 94.3 & 76.2 & 47.4\\
\rowcolor{darkgray!20}
Texture & 11.3 & 18.2 & 28.4 & 55.5 & 94.8 & 76.2 & 47.4\\
Material & 11.0 & 18.2 & 27.7 & 55.9 & 93.6 & 75.4 & 47.0\\
Positation & 11.4 & 18.0 & 28.0 & 55.0 & 93.1 & 75.7 & 46.9\\
Pattern & 10.9 & 16.9 & 28.2 & 55.2 & 94.8 & 76.2 & 47.0\\
Action & 11.2 & 18.1 & 13.9 & 42.2 & 94.9 & 76.4 & 42.8\\
Context & 7.1 & 17.4 & 16.5 & 29.8 & 94.6 & 73.9 & 39.9\\
\bottomrule
\end{tabular}}
% \vspace{-2mm}
\caption{\textbf{Analysis of different prompts.} 
We conduct an ablation study on each visual attribute individually to verify the positive and negative attributes.} \label{tab2_}
\end{table}
\begin{table*}[]
    % \vspace{-3mm}
    \centering
    \resizebox{0.9\textwidth}{!}{
    % \renewcommand\arraystretch{1.3}
    % \setlength{\tabcolsep}{0.9mm}
    % \small
    \begin{tabular}{c|c|c|c|c|c|c|c} \hline
    \toprule
    Methods  & A-847 & PC-459 & A-150 & PC-59 & VOC & VOCb & avg. \\ \hline
    Size & 11.6 & 18.2 & 28.3 & 55.8 & 94.3 & 76.2 & 47.4\\
    Size+Shape & 11.3 & 18.2 & 28.2 & 55.3 & 95.1 & 76.8 & 47.5\\
    Size + Shape +Texture & 11.4 & 18.5 & 28.7 & 54.4 & 95.0 & 77.5 & 47.6\\
    \rowcolor{darkgray!20}
    Size + Shape +Texture + Color & 11.6 & 18.4 & 28.9 & 55.6 & 94.9 & 77.1 & \textbf{47.8}\\
    Size + Shape +Texture + Color + Material & 11.6 & 18.3 & 28.7 & 55.4 & 94.9 & 76.3 & 47.5\\
    Typical attributes & 11.3 & 18.2 & 27.9 & 54.6 & 94.8 & 74.6 & 46.9\\
    \bottomrule
    \end{tabular}}
    \vspace{-2mm}
    \caption{\textbf{Ablation Study on Attribute Combinations.} 
    We conduct an ablation study on different combinations of different attributes and identify the optimal combination.}
    \label{tab2}
    % \end{wraptable}
 \end{table*}
% \end{wraptable}
As the performance may vary across different datasets, we present the average results over all datasets to provide a more robust evaluation of the best approach.
The baseline method, relying on fixed hand-crafted prompts, achieves an average mIoU of 46.8\%. 
Attributes perform differently: color (47.3\%), shape (47.4\%), texture (47.4\%), and size (47.4\%) lead, followed by material (47.0\%), pattern (47.0\%), position (46.9\%), action/state (42.8\%), and contextual relationship (39.9\%). Experimental results indicate that color, size, shape, and texture are the most influential attributes for generating effective descriptive prompts.

Then, we explore different combinations of these attributes in Table~\ref{tab2}. The results show improved average mIoU of 47.5\% (size + shape), 47.6\% (size + shape + texture), 47.8\% (size + shape + texture + color), and 47.5\% (size + shape + texture + color + material), respectively. 
We also query GPT-4 with the prompt: ``\texttt{Describe a \{class name\} with respect to its typical attributes in one sentence. The description must be under 77 tokens as per the CLIP tokenizer.}"
However,  the combinations of well-chosen attributes outperform typical attribute descriptions, highlighting the effectiveness of consistent and strategically selected attributes in enhancing segmentation performance.
 % \vspace{-6mm}
\paragraph{Ablation study for CFM.}
We investigate the impact of the filtered class number in CFM without SAM in Table~\ref{tabk}. 
The results show that performance metrics remain remarkably stable across different $k$ values ranging from 16 to 96. Particularly, stability is observed between $k=32$ and $k=96$, highlighting the model’s robustness to variations in this hyperparameter. While higher $k$ values lead to a slight increase in latency, the corresponding gains in accuracy beyond $k=32$ are marginal. Considering the trade-off between computational efficiency and performance, we therefore select $k=32$ as the optimal default setting. 
% \vspace{-2mm}
\paragraph{Ablation Study for Feature Refinement Module.} 
To evaluate the effectiveness of our feature refinement module, we conduct an ablation study and present the results in Table~\ref{tab4_}. In this experiment, we adopt CLIP ViT-B/16 as the backbone. 
We independently validate the contributions of spatial refinement and class refinement. We observe that integrating both leads to the optimal results, suggesting their complementary roles in enhancing segmentation.
\begin{table}
% \begin{wraptable}{r}{0.6\textwidth} 
    % \vspace{-3mm}
    \renewcommand\arraystretch{1.4}
    \centering
    \resizebox{1\linewidth}{!}{
\begin{tabular}{c|c|c|c|c|c|c|c|c} \hline
\toprule
K  & A-847 & PC-459 & A-150 & PC-59 & VOC & VOCb & avg & Latency (ms) \\ \hline
16 & 11.7 & 19.0 & 30.9 & 58.3 & 95.0 & 79.9 & 49.1 & 339.5\\
\rowcolor{darkgray!20}
32 & 12.8 & 19.7 & 32.1 & 58.0 & 94.8 & 80.0 & 49.6 & \textbf{362.8}\\
% \rowcolor{darkgray!20}
48 & 12.6 & 19.6 & 32.0 & 58.1 & 95.2 & 79.9 & 49.6 & 389.1\\
64& 12.6 & 19.9 & 32.0 & 58.4 & 94.8 & 79.7 & 49.6 & 421.7\\
96 & 12.6 & 19.8 & 32.2 & 58.4 & 94.8 & 79.6 & 49.6 & 460.2\\
\bottomrule
\end{tabular}}
% }
% \vspace{-2mm}
\caption{\textbf{Ablation Study on the Number of Filtered Classes $k$}. } \label{tabk}
 \end{table}
% \end{wraptable}
% \vspace{-9mm}
% \begin{wraptable}{r}{0.6\textwidth} % r=右侧，宽度自己调
\begin{table}
    % \vspace{-2mm} 
    \centering
    % --- 表格 1 ---
    \renewcommand\arraystretch{1.4}
    \resizebox{1\linewidth}{!}{
    % \small 
    \begin{tabular}{c|c|c|c|c|c|c} \hline
    \toprule
    Methods & A-847 & PC-459 & A-150 & PC-59 & VOC & VOCb  \\ \hline
    LSMSeg(w/o FRM) & 11.6 & 18.4 & 28.9 & 55.6 & 94.9 & 77.1 \\
    LSMSeg(w/o Spatial) & 11.6 & 18.5 & 30.5 & 56.8 & 93.2 & 78.6 \\
    LSMSeg(w/o Class) & 11.7 & 19.2 & 30.7 & 57.6 & 95.0 & 78.9 \\
     \rowcolor{darkgray!20}
    LSMSeg(w/ FRM) & \textbf{13.1} & \textbf{20.3} & \textbf{33.2} & \textbf{59.7} & \textbf{95.4} & \textbf{81.1} \\\hline
    % CATSeg$^{\dagger}$ & 11.9 & 18.5 & 31.7 & 57.6 & 94.6 & 77.0 \\
    LSMSeg(w/o SAM) & 12.8 & 19.7 & 32.1 & 58.0 & 94.8 & 80.0 \\
    LSMSeg(w/ Dinov2-B)  & 12.2& 18.8 & 31.1 & 57.2 &  94.9 & 78.7\\ 
    LSMSeg(w/ SAM-B) & 12.5 & 20.3  & 32.1 & 58.8 & 95.2 & 80.2 \\ 
     \rowcolor{darkgray!20}
    LSMSeg(w/ SAM-L)  & \textbf{13.1} & \textbf{20.3} & \textbf{33.2} & \textbf{59.7} & \textbf{95.4} & \textbf{81.1} \\ 
     \bottomrule
    \end{tabular}}
    \vspace{-2mm}
    % \captionsetup{width=0.95\linewidth}
    \caption{\textbf{Ablation study on Feature Refinement Module.} 
We conduct an ablation study to verify the effectiveness of our proposed Feature Refinement Module (FRM).} \label{tab4_}
\end{table}
    % \vspace{2mm}
    % \begin{wraptable}{r}{0.6\textwidth} 
% \begin{table}[h]
% \renewcommand\arraystretch{1.3}
% \setlength{\tabcolsep}{0.05mm}
    % \centering
    
In addition, we investigate the impact of integrating external visual foundation models. Compared to the baseline without additional features, incorporating SAM (both SAM-B and SAM-L) leads to consistent improvements, demonstrating the benefits of leveraging SAM’s strong visual priors. We also evaluate DINOv2-B, which provides moderate gains but remains inferior to SAM-B variants. 
Notably, combining FRM with SAM yields the best overall performance, confirming that these two components are complementary in strengthening the segmentation ability of our model.
This improvement is particularly evident on A-150 and PC-59, where richer prompts and spatial priors strengthen pixel–text alignment for both seen and unseen classes.
\begin{table}
% \end{wraptable}
% \begin{wraptable}{r}{0.6\textwidth}
    % --- 表格 2 ---
    % \vspace{2mm} % 两个表格之间的间距
    % \setlength{\tabcolsep}{1mm}
    \renewcommand\arraystretch{1.4}
    \resizebox{1\linewidth}{!}{
    % \small
    \begin{tabular}{c|c|c|c|c|c|c|c} \hline
    \toprule
    Methods  & A-847 & PC-459 & A-150 & PC-59 & VOC & VOCb & avg. \\ \hline
    mean  & 12.9 & 20.1 & 32.8 & 59.3 & 95.0 & 80.8 & 50.2\\
    concat  & 13.0 & 20.3 & 32.8 & 59.1 & 95.2 & 80.9 & 50.2\\
    \rowcolor{darkgray!20}
    weight generator & \textbf{13.1} & \textbf{20.3} & \textbf{33.2} & \textbf{59.7} & \textbf{95.4} & \textbf{81.1} & \textbf{50.5}\\
    \bottomrule
    \end{tabular}}
    \vspace{-2mm}
    % \captionsetup{width=0.95\linewidth}
    \caption{\textbf{Ablation study on different fusion strategies.} Here, `mean' denotes the element-wise average of CLIP and SAM features, while `concat' refers to feature concatenation followed by a projection for dimensional alignment.} \label{tab3}
    \end{table}
We additionally present an ablation study on visual feature fusion strategies in Table~\ref{tab3}.
Compared with simple averaging (mean) and straightforward concatenation (concat), our proposed weight generator achieves the best performance with an average score of 50.5\%. This demonstrates that adaptively learning fusion weights provides a more effective balance between CLIP and SAM features, leading to consistently improved results across different datasets.
% \vspace{-2mm}
\paragraph{Model efficiency.} 
In Table~\ref{tab:efficiency_comparison}, we compare the efficiency of LSMSeg with recent methods in terms of learnable parameters, training time, inference latency, and computational cost (GFLOPs). 
LSMSeg demonstrates strong efficiency in both training and inference, benefiting from the category filtering module.
When combined with SAM-L, LSMSeg maintains competitive efficiency while further improving segmentation performance, highlighting the effectiveness of integrating strong visual information with our lightweight design.
\begin{table}
    \renewcommand\arraystretch{1.4}
    \resizebox{1\linewidth}{!}{
    % \small
    \begin{tabular}{c|cccc}
        \hline
        Methods & Learnable Params (M) & Training Time (min) & Latency (ms)&  GFLOPs \\ \hline
        ZegFormer & 103.3 & 1148.3 & 2700 &  19,425.6\\ 
        %  ZSSeg & 25.12 & 28.96 \\ \hline
         OVSeg & 102.8 & - & 2000 &  19,345.6\\ 
         CAT-Seg& 70.3 & 693.7 & 535.2 &  3459.7\\
         % SED & 180.8 & 358.9 & 1702 & 1068.9\\
         \rowcolor{darkgray!20}
         LSMSeg(w/o SAM-L) & \textbf{70.3} & \textbf{446} & \textbf{362.8} &  \textbf{2122.0}\\ 
         LSMSegLSMSeg(w/ SAM-L) & \underline{73.4} & \underline{546} & \underline{426.0} & \underline{3140.6}\\ \hline
    \end{tabular}}
    \vspace{-2mm}
    \caption{\textbf{Efficiency comparison}. All results are measured with the Nvidia-L40 GPU.}
    \label{tab:efficiency_comparison}
% \end{table}
\end{table}
\section{Conclusion}
In this work, we introduce LSMSeg, a novel framework that advances open-vocabulary semantic segmentation (OVSS) by effectively modeling the relationship between textual and visual representations. 
By leveraging GPT-4 to generate attribute-based text prompts, LSMSeg enriches the semantic content of textual inputs, enabling more precise pixel-level alignment with visual features. 
Furthermore, our Category Filtering Module (CFM) and Feature Refinement Module optimize computational efficiency and segmentation accuracy by pruning irrelevant categories.
Lastly, we propose a Feature Refinement Module that dynamically integrates with CLIP visual features to achieve both spatial and class-level feature refinement.
Extensive experiments show that LSMSeg not only achieves state-of-the-art performance but also maintains efficient inference with lower latency.

% \section{Final copy}

% You must include your signed IEEE copyright release form when you submit your finished paper.
% We MUST have this form before your paper can be published in the proceedings.

% Please direct any questions to the production editor in charge of these proceedings at the IEEE Computer Society Press:
% \url{https://www.computer.org/about/contact}.
% \clearpage

{
    \small
    \bibliographystyle{ieeenat_fullname}
    \bibliography{main}
}


\end{document}